\documentclass[runningheads]{llncs}
\usepackage[T1]{fontenc}
\usepackage{xcolor}
\usepackage{graphicx}
\usepackage{booktabs} 
\usepackage{array} 
\usepackage{multirow} 
\usepackage{pifont}
\usepackage{amsmath}
\usepackage{amssymb}
\usepackage{comment}
\usepackage{tabularx}
\begin{document}
\title{EmCom-Diffusion: Probing Visual Reflection in Emergent Languages via Image Generation}
\titlerunning{EmCom-Diffusion: Probing Visual Reflection}
%
%
\author{Haruumi Omoto\inst{1}\orcidID{0009-0004-3905-6412} \and
Tadahiro Taniguchi\inst{1,2}\orcidID{0000-0002-5682-2076}}
\authorrunning{H. Omoto and T. Taniguchi}
%
\institute{Graduate School of Informatics, Kyoto University, Kyoto, Japan \and
Research Organization of Science and Technology, Ritsumeikan University, Shiga, Japan\\
\email{omoto.haruumi.45u@st.kyoto-u.ac.jp},
\email{taniguchi@i.kyoto-u.ac.jp}}
\maketitle              
\begin{abstract}
Measuring the extent to which emergent languages encode the visual content of their inputs is an open problem. We refer to this property as \textbf{visual reflection}: the extent to which emergent messages preserve information about their source images that can be recovered without appeal to the speaker--listener pair that produced them. Existing metrics measure it only indirectly, through proxies such as human-defined concept inventories, natural-language captions, structural distance correlations, or Referential Game accuracy, each of which can either miss visual content the message encodes or credit content it does not. We propose \textbf{EmCom-Diffusion}, an evaluation framework that measures visual reflection directly: it reconstructs each input image from its emergent message and compares the reconstruction with the original image itself, rather than with human-defined targets. Concretely, it fine-tunes a pretrained text-to-image diffusion model on (image, emergent-message) pairs and scores visual reflection as the perceptual similarity between the reconstructed and original images, operating generatively rather than discriminatively. Instantiating it on MS-COCO with a Referential Game, we validate the metric against random and fixed-token baselines under three pretrained visual encoders, and compare it against four existing metrics (CBM, supervised translation, TopSim, and R@1). EmCom-Diffusion captures visual content the other metrics miss. Our code is available at \url{https://github.com/Tanichu-Laboratory/EmCom-Diffusion}.

\end{abstract}
\section{Introduction}

\begin{table}[!b]
\centering
\footnotesize
\caption{Existing emergent-language evaluations measure visual reflection through different proxies. EmCom-Diffusion uniquely compares emergent messages with the input images they describe.}
\label{tab:eval-comparison}
\setlength{\tabcolsep}{5pt}
\begin{tabular}{lll}
\toprule
Method & What it compares & Reflection aspect captured \\
\midrule
CBM         & Tokens $\leftrightarrow$ concepts        & Predefined-concept alignment \\
Translation & Messages $\leftrightarrow$ captions      & Natural-language translatability \\
TopSim      & Distances $\leftrightarrow$ distances    & Distance preservation \\
R@1         & Messages $\leftrightarrow$ outcomes      & Discriminative communicability \\
\textbf{EmCom-Diffusion} & \textbf{Messages $\leftrightarrow$ images} & \textbf{Generative reconstructability} \\
\bottomrule
\end{tabular}
\end{table}

Emergent Communication (EmCom) is a research framework in which agents develop an emergent language (EL) from scratch through interaction~\cite{lazaridou2020emergentsurvey}; a central open problem is how to measure what this language encodes about its visual inputs. We call this property \textbf{visual reflection}, and understanding it is important for interpreting what an emergent language means. It also bears on a broader question: why language comes to encode the structure of the physical world~\cite{HarrisDistributionalSemanticsOrigin,taniguchiCWM}. LLMs trained only on text appear to capture aspects of this physical structure~\cite{LLMSpaceTime,LLaVA,LLMseehear}, but this evidence comes from mature human language that has already converged on such encodings. Emergent communication instead lets us observe how this structure comes to be encoded as a language forms from scratch. Within EmCom, however, attention has largely gone to the formal properties of the resulting language, such as the compositionality, generalization, and syntactic structure of its token sequences~\cite{chaabouni-etal-2020-compositionality,mu2021emergent,resnick2020cap}, while visual reflection has received comparatively little attention. The few studies that do examine it report that emergent languages can drift from their visual input despite high task success~\cite{howagentSee,CuriousCaseEmCom}. We therefore focus on visual reflection in emergent languages.

However, the lack of established evaluation metrics has limited systematic investigation of visual reflection. Metrics for evaluating semantic content in emergent languages fall into three families~\cite{Peters_2025emcomsurvey}. (i) Reference-based metrics compare emergent tokens against human-defined references~\cite{emcomtranslation,mu2021emergent}. Concept-Best-Matching (CBM)~\cite{carmeli2024concept} matches tokens to predefined oracle concepts via bipartite matching; supervised translation methods~\cite{yao2022linking} train translators on paired EL--NL data and report translation quality.
(ii) Structural correspondence metrics rely on structural alignment between message and input spaces; the most common is topographic similarity (TopSim)~\cite{kirby2006topsim}, which computes the Spearman correlation between pairwise distances in the input space and the message space.
(iii) Task-performance metrics are based on the agents' performance on the communication task itself; the representative example is Referential Game accuracy (R@1)~\cite{lazaridou2017referential}, which assesses whether the listener can identify the target image given the emergent message.
Formal definitions of the baselines we use (CBM, Translation, TopSim, R@1) are given in §\ref{sec:comparison}. 
Each family is inherently limited by its specific proxy for visual reflection (Table~\ref{tab:eval-comparison}). 
Reference-based metrics miss any visual content that falls outside a predefined concept inventory. 
Structural metrics treat distance correlation as the sole indicator of visual reflection; consequently, messages that encode visual content without preserving distances are incorrectly evaluated as lacking visual information. 
Task-performance metrics conflate communicative success with the actual presence of visual content. Indeed, an emergent language can contribute to successful target identification while failing to sufficiently reflect the visual content of the input~\cite{CuriousCaseEmCom}. 
Ultimately, no existing method directly compares emergent messages with the input images themselves. 

To address the lack of metrics that directly measure visual reflection, we propose \textbf{EmCom-Diffusion}. EmCom-Diffusion consists of three steps (Fig.~\ref{fig:overview}): (a) obtaining paired data of image inputs and emergent-language messages from a communication game; (b) fine-tuning a pretrained text-to-image diffusion \allowbreak model~\cite{stablediffusion} on this paired data; (c) evaluating the generated images using metrics such as CLIPScore~\cite{clipscore}. This method can directly measure visual reflection by comparing the original and generated images. EmCom-Diffusion has two key advantages: it is annotation-free, requiring no predefined concept inventories or captions, and it is generative, producing the image from the message alone and comparing it directly with the original, so that only the visual content encoded in the message can appear in the generated image, allowing visual reflection to be measured faithfully. To verify that EmCom-Diffusion serves as a useful metric for measuring visual reflection in emergent languages, we instantiate this pipeline with a Referential Game~\cite{chaabouni2022emergentScale,lazaridou2017referential} on MS-COCO~\cite{microsoftCOCO} and conduct two evaluations. First, we compare EmCom-Diffusion scores of emergent languages against baselines such as natural-language captions and random tokens, showing that visual reflection can indeed be evaluated by EmCom-Diffusion. Second, we compare EmCom-Diffusion against existing metrics (CBM, Translation, TopSim, and R@1) and show that EmCom-Diffusion captures aspects of visual reflection that these metrics cannot.

The contributions of this paper are as follows.
\begin{itemize}
\item We propose EmCom-Diffusion, a novel metric that reconstructs the input image from the emergent-language message, and show that it directly evaluates visual reflection in emergent languages.
\item We show that EmCom-Diffusion captures aspects of visual reflection in emergent languages that existing metrics have overlooked.
\end{itemize}

\section{Methods}
\label{sec:methods}
\subsection{Visual Reflection}
\label{sec:vr}

\begin{figure}[t]
    \centering
    \includegraphics[width=\linewidth]{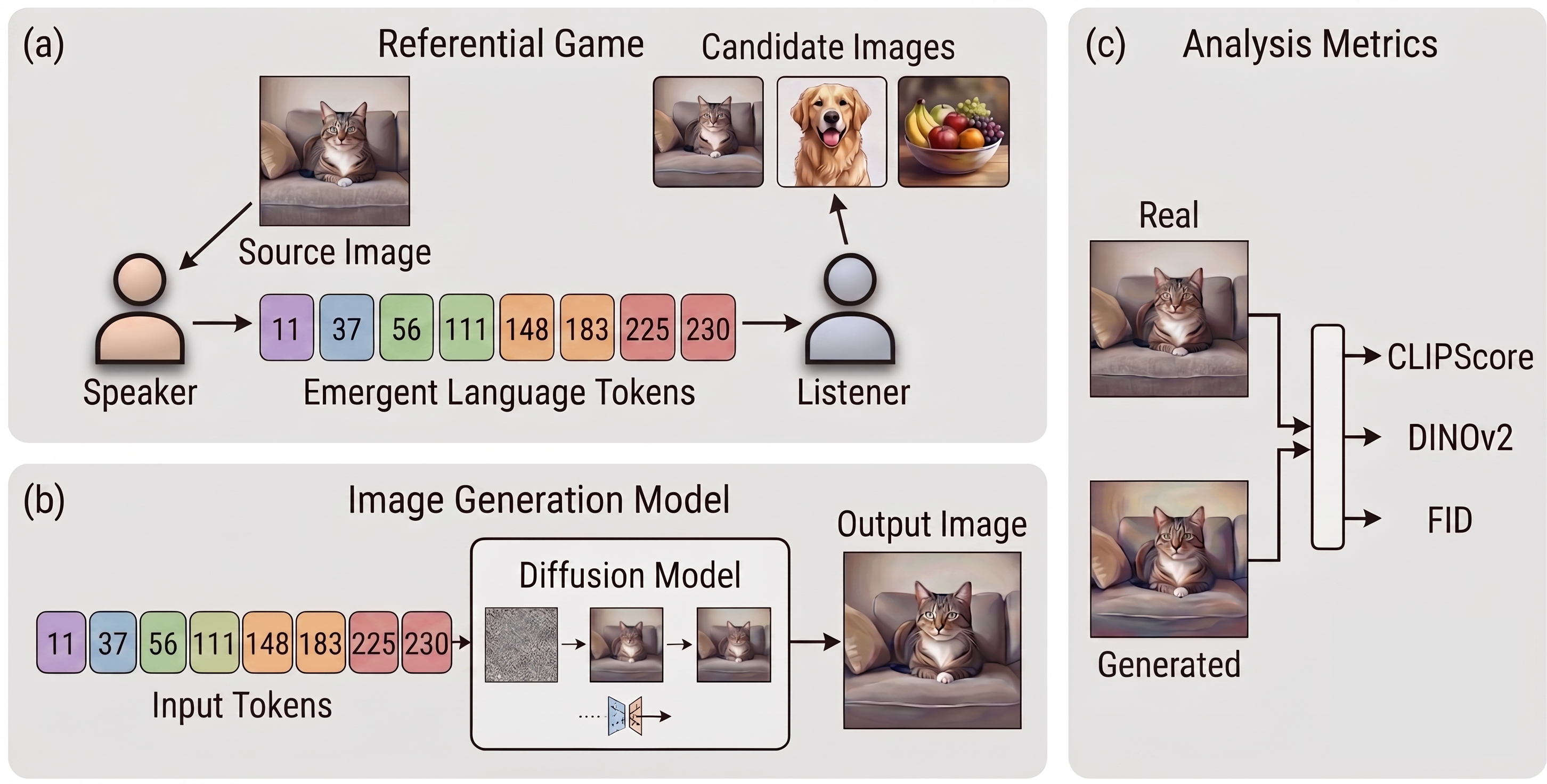}
    \caption{Overview of our EmCom-Diffusion framework. (a) Agents play a communication game (here, a Referential Game) on images, producing emergent languages as sequences of discrete tokens. (b) We fine-tune a pretrained image generation model on pairs of images and emergent languages. (c) Generated images are compared with the original images using a suite of evaluation metrics to assess what visual information emergent languages reflect.}
    \label{fig:overview}
\end{figure}
 
We say that an emergent language reflects the visual content of an image when its messages encode information about that content in a way that can be recovered without further appeal to the speaker--listener pair that produced them. Formally, given a corpus $\mathcal{D} = \{(x_i, m_i)\}_{i=1}^{N}$ of (image, message) pairs, an emergent language has high visual reflection if some decoder, independent of the original speaker--listener system, can map messages back to the visual content of their corresponding images. This is the property our framework is designed to measure.
 
\subsection{The Proposed Framework: EmCom-Diffusion}
\label{sec:framework}
 
\paragraph{Central idea.}
Fig.~\ref{fig:overview} gives an overview of the framework. Given $\mathcal{D}$, we fine-tune a text-conditional diffusion model~\cite{stablediffusion} $G_\theta$ to reconstruct each image $x$ from its emergent message $m$, then quantify reflection by the perceptual similarity between $\hat{x} = G_\theta(m)$ and $x$, averaged over a held-out set. The diffusion model is a learned external decoder whose ability to recover visual content from messages is our measurement instrument.
 
\paragraph{Formal definition.}
Let $\mathcal{L}_{\mathrm{diff}}(x, m;\theta)$ denote the standard latent-diffusion
denoising loss for image $x$ conditioned on message $m$.
We fine-tune the model on the emergent corpus by minimizing it,
\begin{equation}
  \theta^{*} = \arg\min_{\theta}\;
  \mathbb{E}_{(x,m)\sim\mathcal{D}_{\mathrm{train}}}
  \big[\mathcal{L}_{\mathrm{diff}}(x, m;\theta)\big].
\end{equation}
At evaluation, we sample a reconstruction $\hat{x}=G_{\theta^{*}}(m)$ from the
fine-tuned model and, given a perceptual similarity function $s$ and an evaluation
set $\mathcal{D}_{\mathrm{eval}}=\{(x_i,m_i)\}_{i=1}^{N}$, define the
EmCom-Diffusion score as
\begin{equation}
\label{eq:em-diff-score}
  \mathrm{EmCom\text{-}Diffusion}(\mathcal{D}_{\mathrm{eval}};\,s)
  = \frac{1}{N}\sum_{i=1}^{N} s\big(G_{\theta^{*}}(m_i),\,x_i\big).
\end{equation}
 
\paragraph{Why this measures visual reflection.}
Two properties make Eq.~\eqref{eq:em-diff-score} a faithful measurement of visual reflection.
 
\textbf{(i) No human-defined ground truth.}
The standard of comparison in Eq.~\eqref{eq:em-diff-score} is the image $x$ itself. We do not impose any human-curated annotation, such as a fixed concept inventory or paired natural-language captions, as the target for evaluation. This is crucial because the visual structure encoded by an emergent language need not align with human-natural concepts: a token may consistently track a visual regularity that does not correspond to any English noun or annotated category. Such regularities are invisible to evaluators built on human annotations but remain measurable in our framework, as the diffusion model can in principle learn any token-to-visual mapping supported by the training pairs.
 
\textbf{(ii) Generative rather than discriminative evaluation.}
The core mechanism of EmCom-Diffusion is \emph{generative}. Because $G_{\theta^{\ast}}$ is forced to produce a complete image from the message $m$ alone, in principle, any visual content absent from $m$ will also be absent from the generated image $\hat{x}$, resulting in a lower similarity score with $x$ under $s$. While this ideal is subject to practical nuances, such as the diffusion model hallucinating plausible details from an impoverished $m$ or the similarity function $s$ introducing its own biases (e.g., CLIP's preference for salient foreground objects), the generative bottleneck fundamentally restricts information hiding. By contrast, a natural alternative is to use a \emph{discriminative} approach, such as computing the similarity between encoded messages and encoded images. Discriminative evaluations compress $m$ and $x$ into low-dimensional embeddings, creating a degree of freedom on the evaluator side: the encoders can quietly discard information missing from $m$ while artificially maintaining high alignment. Generation removes this degree of freedom, making it a more faithful instrument for measuring visual reflection.

\subsection{Comparison Methods}
\label{sec:comparison}
 
We compare EmCom-Diffusion against four existing evaluations of emergent languages. For each, we state the definition and identify how its score can deviate from visual reflection; implementation details (datasets, hyperparameters, evaluation sets) are reported in Appendix~\ref{app:comparison}.
 
\textbf{CBM}~\cite{carmeli2024concept}.
CBM aligns each token $v \in \mathcal{V}$ to at most one concept from a predefined inventory $\mathcal{C}$ by maximum-weight bipartite matching. The edge weight $q_{vc}$ represents the co-occurrence frequency between $v$ and an annotated concept $c$ across the corpus. The metric then reports the normalized total weight of the optimal matching. Because $q_{vc}$ is defined only over $\mathcal{C}$, visual content that the message encodes outside the inventory contributes zero to every edge weight and cannot be reflected in the score, regardless of how faithfully it is encoded.

\textbf{Supervised translation}~\cite{yao2022linking}.
A seq2seq model $f_\theta : \mathcal{V}^K \to \mathrm{NL}$ is fit on paired $(m_i, c_i)$ data, where $c_i$ is a natural-language caption of image $x_i$, and translation quality (e.g., ROUGE-L~\cite{lin-och-2004-automaticROUGE-L}) is reported on held-out pairs. The score is strictly bounded by the contents of the reference captions $\{c_i\}$. Visual content encoded in $m_i$ but absent from $c_i$ receives no credit.

\textbf{TopSim}~\cite{kirby2006topsim}.
Let $d_{\mathrm{msg}}$ be the normalized edit distance between messages and $d_{\mathrm{input}}(x_i, x_j) = 1 - \cos(\phi(x_i), \phi(x_j))$ for a pretrained image encoder $\phi$. TopSim is the Spearman rank correlation $\rho_s$ between $\{d_{\mathrm{msg}}(m_i, m_j)\}_{i<j}$ and $\{d_{\mathrm{input}}(x_i, x_j)\}_{i<j}$. The score depends only on the rank order between the two arrays. A language whose mapping $x \mapsto m$ perfectly reflects visual content via a code that does not preserve rank order (e.g., a bijective hash assigning visually similar images to message pairs at arbitrary $d_{\mathrm{msg}}$) yields a TopSim score near zero despite faithful reflection.

\textbf{R@1}~\cite{lazaridou2017referential}.
For each evaluation example, one target image $x_i^+$ and a set of distractor images are drawn uniformly at random. The listener selects $\hat{x}_i = \arg\max_x s(m_i, x)$ using its learned scoring function $s$, and R@1 is the empirical accuracy $\frac{1}{N}\sum_i \mathbf{1}[\hat{x}_i = x_i^+]$. However, R@1 is highly sensitive to the evaluation setup, including whether distractors are drawn uniformly at random or as visually similar hard negatives, as well as how many distractors are used, such that the same message and the same listener can yield substantially different scores. R@1 therefore reflects properties of the evaluation setup, such as the distractor distribution, rather than the visual content encoded in the message itself, and cannot stably measure visual reflection.

EmCom-Diffusion places emergent messages directly in correspondence with the input images they describe, removing the predefined inventories, natural-language proxies, structural assumptions, and discriminative shortcuts that limit the four methods above.

\section{Experiments}\label{sec:experiments}

\subsection{Purpose and Common Setup}
\label{subsec:setup}

We conduct two experiments to evaluate EmCom-Diffusion as a metric for visual reflection in emergent languages. \textbf{Experiment~1} verifies that EmCom-Diffusion can effectively measure visual reflection by comparing emergent-language scores against baselines with known information content. \textbf{Experiment~2} compares EmCom-Diffusion against existing metrics to demonstrate that it captures aspects of visual reflection that other methods cannot.

\paragraph{Dataset and emergent languages.}
We use the MS-COCO 2017 dataset~\cite{microsoftCOCO} as the source of natural images. To obtain emergent languages, we train a speaker--listener pair from scratch via a \emph{Referential Game}~\cite{chaabouni2022emergentScale,lazaridou2017referential}: the speaker observes a target image and emits a discrete message that the listener uses to identify the target from a set of distractor images. This setup encourages the emergent language to encode visual information that discriminates among images. After training, only the speaker is used to produce one emergent-language message per image. All metrics are evaluated on a fixed held-out subset of $5{,}000$ images from the MS-COCO 2017 validation set. Speaker architecture, vocabulary settings, and training details are reported in Appendix~\ref{app:speaker}.

\paragraph{Evaluation metrics.}
We evaluate EmCom-Diffusion along three axes:

\textbf{Similarity to the original image.} This is the EmCom-Diffusion score itself: the perceptual similarity $s
$ of Eq.~\eqref{eq:em-diff-score}, here instantiated as the cosine similarity between embeddings of the original and generated images. We compute it with three encoders spanning different supervision paradigms: \textbf{CLIP-img}~\cite{clip} (vision--language contrastive learning), \textbf{DINOv2}~\cite{oquab2024dinov2} (self-supervised vision-only), and \textbf{SigLIP}~\cite{zhai2023sigmoid} (sigmoid loss vision--language). Multiple encoders ensure that our findings are robust to encoder choice.

\textbf{Alignment with natural-language description.} \textbf{CLIP-text}~\cite{clip}: cosine similarity between the CLIP text embedding of the original caption and the CLIP image embedding of the generated image. This metric provides a secondary axis: while it does not directly measure visual reflection per §\ref{sec:vr}, it captures whether the generation aligns with the natural-language description of the original image.

\textbf{Quality of the generated image distribution.} \textbf{FID}~\cite{FID}: distance between generated and original image distributions in Inception feature space; \textbf{Vendi}~\cite{friedman2023vendi}: effective diversity of the generated image distribution; \textbf{Recall}~\cite{kynkaanniemi2019improved}: coverage of the original image distribution by the generated one.

\begin{table}[!b]
\centering
\small
\setlength{\tabcolsep}{3pt}
\caption{Quantitative comparison across reference inputs under the default setting (K{=}8, V{=}256). $\uparrow$ $\downarrow$ indicates higher/lower is better. Per-image metrics are averaged over n=5{,}000 pairs; distribution metrics are computed on the full set. †  SD-NL uses privileged human-written captions; it is a strong reference EL is not expected to surpass.}
\resizebox{\linewidth}{!}{%
\begin{tabular}{l ccc c ccc}
\toprule
 & \multicolumn{3}{c}{Similarity to original image $\uparrow$} & \begin{tabular}[c]{@{}c@{}}Alignment\\with NL\end{tabular} & \multicolumn{3}{c}{Distribution} \\
\cmidrule(lr){2-4} \cmidrule(lr){5-5} \cmidrule(lr){6-8}
Input & CLIP-img & DINOv2 & SigLIP & CLIP-text $\uparrow$ & FID $\downarrow$ & Vendi $\uparrow$ & Recall $\uparrow$ \\
\midrule
SD-NL$^\dagger$ & 0.726 & 0.438 & 0.730 & 0.313 & 25.61 & 22.4 & 0.345 \\
\textbf{EL}     & \textbf{0.638} & \textbf{0.247} & \textbf{0.642} & \textbf{0.217} & \textbf{29.01} & \textbf{20.7} & \textbf{0.255} \\
Random          & 0.485 & 0.022 & 0.493 & 0.153 & 50.19 & 17.4 & 0.227 \\
Fixed           & 0.492 & 0.031 & 0.499 & 0.153 & 115.75 & 7.3 & 0.015 \\
\bottomrule
\end{tabular}%
}
\label{tab:main_results}
\end{table}

\subsection{Experiment 1: Validity}
\label{subsec:validity}

\paragraph{Setup.}
To verify that EmCom-Diffusion serves as a valid metric for visual reflection, we evaluate four input conditions with known relative information content:

\textbf{Random tokens}: token sequences sampled uniformly from the emergent vocabulary. \textbf{Fixed tokens}: a single token sequence used for all images. Both carry no image-specific information.

\textbf{Emergent Language (EL)}: the trained emergent language, which encodes discriminative visual information through the Referential game.

\textbf{SD-NL}: Stable Diffusion fine-tuned on caption-image pairs, where each input is one of the five MS-COCO captions of the original image. We treat it as a strong reference rather than a fair comparison: it relies on privileged human-written captions and the pretrained generator is already aligned to natural language, so EL is not expected to match it. 

A valid metric should rank these as $\textbf{Random/Fixed} < \textbf{EL} < \textbf{SD-NL}$: EL carries visual information learned through the Referential game, while SD-NL additionally draws on privileged human captions and the generator's existing alignment to natural language.

\begin{figure}[t]
\centering
\setlength{\tabcolsep}{2pt}
\renewcommand{\arraystretch}{1.1}
\begin{tabular}{p{0.18\linewidth}p{0.18\linewidth}p{0.18\linewidth}p{0.18\linewidth}p{0.18\linewidth}}
 
\toprule
\centering Original &
\centering SD-NL &
\centering EL \newline ($K{=}8,V{=}256$) &
\centering EL \newline ($K{=}16,V{=}64$) &
\centering\arraybackslash Random \\
\midrule
\addlinespace[2pt]
 
\centering\scriptsize\textit{``There is an elephant standing near a tree.''} &
\centering\scriptsize\textit{(same as Original)} &
\centering\scriptsize\texttt{[226,226,226,226, 226,226,186,230]} &
\centering\scriptsize\texttt{[6,6,6,6,6,6,6,6, 6,6,6,6,6,19,6,6]} &
\centering\arraybackslash\scriptsize\texttt{[223,248,117,161, 223,84,201,170]} \\
\includegraphics[width=\linewidth]{figures/qualitative/ex1_orig.jpg} &
\includegraphics[width=\linewidth]{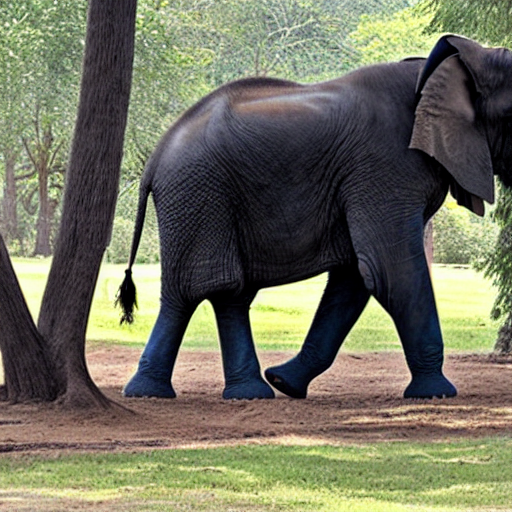}&
\includegraphics[width=\linewidth]{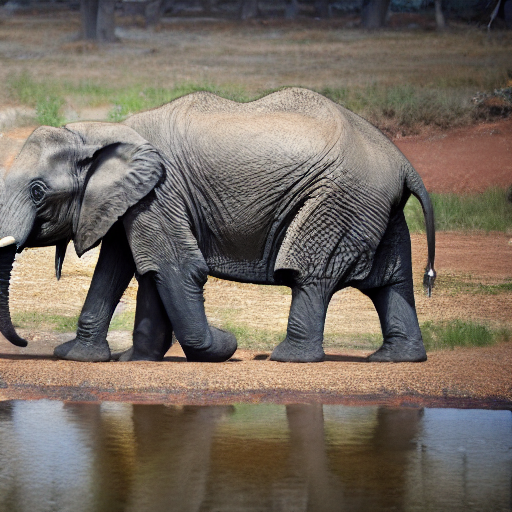} &
\includegraphics[width=\linewidth]{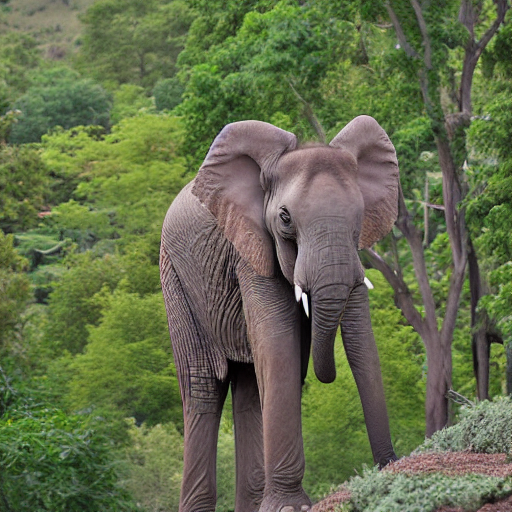} &
\includegraphics[width=\linewidth]{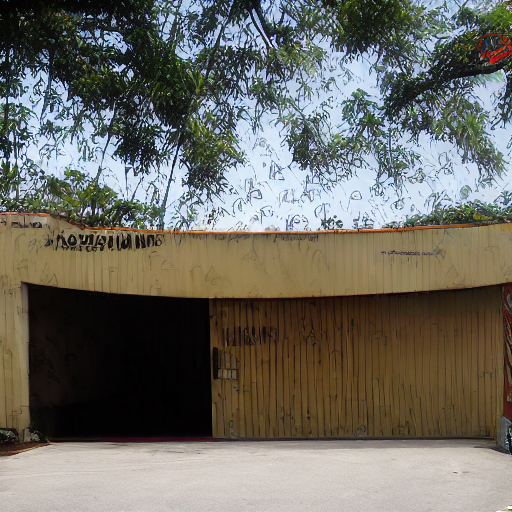} \\
\centering\scriptsize{--} &
\centering\scriptsize CLIP-img: 0.889 &
\centering\scriptsize CLIP-img: 0.904 &
\centering\scriptsize CLIP-img: 0.919 &
\centering\arraybackslash\scriptsize CLIP-img: 0.543 \\
\addlinespace[0.7em]
 
\centering\scriptsize\textit{``Two surfers catching two different waves, one after the other.''} &
\centering\scriptsize\textit{(same as Original)} &
\centering\scriptsize\texttt{[96,96,96,96, 96,96,194,72]} &
\centering\scriptsize\texttt{[13,13,13,44,13, 13,13,13,13,13, 13,13,13,44,13,13]} &
\centering\arraybackslash\scriptsize\texttt{[206,67,87,247, 235,223,207,47]} \\
\includegraphics[width=\linewidth]{figures/qualitative/ex2_orig.jpg} &
\includegraphics[width=\linewidth]{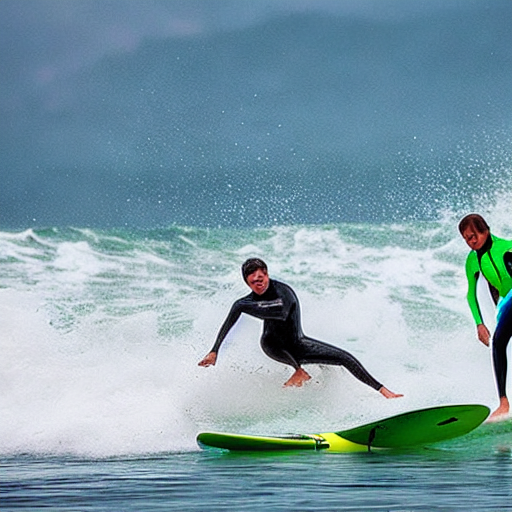} &
\includegraphics[width=\linewidth]{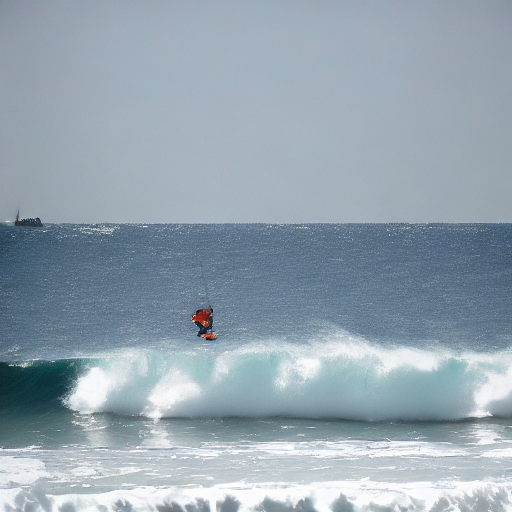} &
\includegraphics[width=\linewidth]{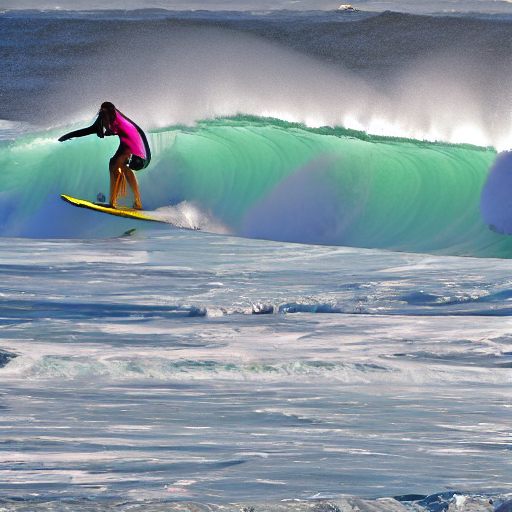} &
\includegraphics[width=\linewidth]{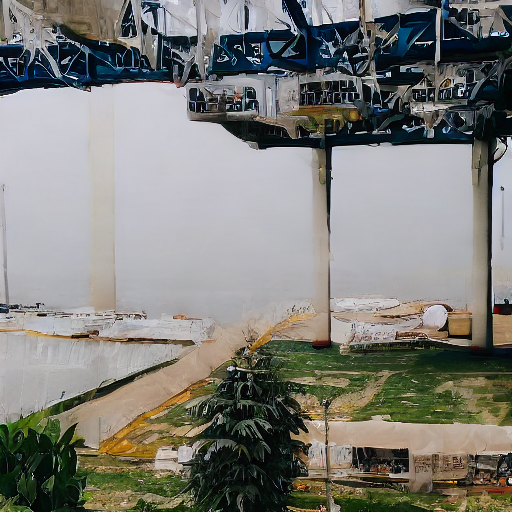} \\
\centering\scriptsize{--} &
\centering\scriptsize CLIP-img: 0.850 &
\centering\scriptsize CLIP-img: 0.900 &
\centering\scriptsize CLIP-img: 0.914 &
\centering\arraybackslash\scriptsize CLIP-img: 0.614 \\
\addlinespace[0.7em]

\end{tabular}
\caption{Qualitative comparison. CLIP-img cosine similarity to the original is given below each generated image (omitted for the Original column, trivially $1.0$). The two EL columns ($K{=}8, V{=}256$ [default] and $K{=}16, V{=}64$) illustrate that the qualitative behavior is not specific to a single capacity setting. SD-NL outperforms EL on average (Table~\ref{tab:main_results}); the examples here reflect individual sample variation rather than the aggregate trend.}
\label{fig:qualitative}
\end{figure}

\paragraph{Results}
Table~\ref{tab:main_results} reports the seven metrics under the default setting. The expected ordering $\textbf{Random/Fixed} < \textbf{EL} < \textbf{SD-NL}$ is observed across all metrics, confirming that EmCom-Diffusion correctly ranks input conditions by their information content. The gap is largest on DINOv2 (EL $0.247$ vs.\ random $0.022$), a vision-only encoder trained without any language supervision; this shows that the metric detects visual content beyond mere distributional priors of MS-COCO. Fig.~\ref{fig:qualitative} confirms this qualitatively: random tokens generate images with little visual correspondence to the originals, whereas emergent languages generate images that visibly resemble them.

\begin{table}[!b]
\centering
\small
\caption{Multi-seed analysis with $8$ seeds per input, averaged over $1{,}000$ samples, in CLIP image feature space. $^\dagger$\, SD-NL uses privileged human captions and is a strong reference rather than a fair comparison.}
\label{tab:multiseed}
\setlength{\tabcolsep}{4pt}
\begin{tabular}{lccc}
\toprule
Input & Intra-token $\downarrow$ & Inter-token $\uparrow$ & Distance to GT $\downarrow$ \\
\midrule
SD-NL $^\dagger$ & 0.176 & 0.398 & 0.215 \\
\textbf{EL} & \textbf{0.259} & \textbf{0.333} & \textbf{0.273} \\
Random & 0.312 & 0.224 & 0.427 \\
Fixed & 0.327 & 0.000 & 0.429 \\
\bottomrule
\end{tabular}
\end{table}

\paragraph{Multi-seed analysis.}
Because the diffusion model is stochastic, a meaningful probe must measure input-specific signal rather than artifacts of sampling. We conduct a multi-seed analysis: for 1{,}000 sampled images, we generate 8 images per token sequence using different random seeds and measure three quantities in CLIP image feature space (Table~\ref{tab:multiseed}): \textbf{intra-token distance} (variability across seeds from the same input), \textbf{inter-token distance} (separation across different inputs), and \textbf{distance to ground truth} (distance from generated images to the original). The expected ordering is preserved on all three metrics, confirming that EmCom-Diffusion reliably distinguishes input-specific signal from stochastic sampling noise.

\subsection{Experiment 2: Comparison with Existing Metrics}
\label{subsec:comparison}
\paragraph{Setup.} We conduct three targeted comparisons, each isolating the structural limitation identified for a representative baseline in §\ref{sec:comparison}: (a) supervised translation~\cite{yao2022linking}; (b) CBM~\cite{carmeli2024concept} and TopSim~\cite{kirby2006topsim}; (c) R@1~\cite{lazaridou2017referential}. Each comparison constructs an image subset that exposes the targeted limitation. We define two images as \textit{visually similar} if their CLIP ViT-L/14 cosine similarity exceeds $\tau = 0.7$. This is the $99$th percentile of pairwise similarities in our evaluation set, a deliberately strict threshold: only the top $\sim 1\%$ of pairs qualify, comparable in selectivity to requiring two images to share their dominant object category ($\approx 1/80 = 1.25\%$). In every comparison, EmCom-Diffusion scores a candidate by the DINOv2 cosine similarity between the candidate image and the reconstruction $\hat{x}_A = G_{\theta^{*}}(m_A)$ generated from the anchor's message, and selects the higher-scoring candidate. The DINOv2 readout is distinct from the CLIP
ViT-L/14 encoder used to define visual similarity, so identification does not reduce to the criterion that defines the triplets.

\paragraph{(a) EmCom-Diffusion captures visual distinctions where
translation-based evaluation cannot.}

\begin{table}[!b]
\centering
\small
\caption{Triplet identification accuracy (\%) under varying caption-similarity control $\varepsilon$ ($N=10{,}000$ triplets per row; brackets are 95\% bootstrap CIs, $10{,}000$ resamples, percentile method, seed 42). As $\varepsilon$ decreases, caption-level signal is increasingly suppressed: translation evaluation degrades by $18$ pp while EmCom-Diffusion by only $8$ pp; the bottom row reports this drop (from $\varepsilon{=}0.20$ to $0.02$) in pp.}
\begin{tabular}{lcc}
\toprule
$\varepsilon$ & EmCom-Diffusion (Ours) &Translation \\
\midrule
$0.20$ (loose) &70.8 [ 69.9, 71.7 ] & 73.6 [ 72.8, 74.5 ] \\
$0.10$         &65.5 [ 64.6, 66.4 ] & 63.7 [ 62.7, 64.6 ] \\
$0.05$         &63.5 [ 62.6, 64.5 ] & 57.1 [ 56.2, 58.1 ] \\
$0.02$ (tight) &63.3 [ 62.3, 64.2 ] & 55.4 [ 54.5, 56.4 ]  \\
\midrule
Drop & \textbf{-7.6 [ -8.9, -6.2 ]} & -18.2 [ -19.5, -16.9 ]  \\
\bottomrule
\end{tabular}
\label{tab:translation_eps}
\end{table}

Translation evaluation is bounded by GT caption content (§\ref{sec:comparison}); we test whether this bound limits its discrimination when caption similarity is held constant. Defining
caption similarity $\mathrm{cap\_sim}$ between two images as the cosine similarity of their average CLIP text embeddings over the five MS-COCO captions, we find that high $\mathrm{cap\_sim}$ rarely entails visual similarity: $96.7\%$ of pairs with $\mathrm{cap\_sim} \geq 0.50$ are visually dissimilar, $89.5\%$ at $\geq 0.60$, and $72.1\%$ at $\geq 0.70$.

For each anchor $A$, candidates $B$ (visually similar) and $C$ (dissimilar) are sampled under $|\mathrm{cap\_sim}(A, B) - \mathrm{cap\_sim}(A, C)| \leq \varepsilon$, suppressing caption-level signal. The translation score is the CLIP cross-modal similarity between $m_A$'s translation and the candidate's GT captions.

Table~\ref{tab:translation_eps} reports the results. At $\varepsilon = 0.20$, translation performs comparably to EmCom-Diffusion ($73.6\%$ vs $70.8\%$). As $\varepsilon$ decreases to $0.02$, translation degrades to $55.4\%$ ($-18.2$ pp), while EmCom-Diffusion retains $63.3\%$ ($-7.6$ pp). The asymmetric drop confirms that translation's signal is largely caption-derived.

\begin{table}[!b]
\centering
\small
\caption{Triplet identification accuracy under matched edit distance. Chance is $50.0\%$; TopSim achieves exactly chance by construction.}
\begin{tabular}{lcccc}
\toprule
Setting & $N$ & EmCom-Diffusion (Ours) & TopSim & CBM \\
\midrule
Edit distance $1$ & $880$  & $\mathbf{77.7\%}$ & $50.0\%$ & $68.8\%$ \\
Edit distance $2$ & $5{,}808$ & $\mathbf{78.0\%}$ & $50.0\%$ & $62.8\%$ \\
Edit distance $3$ & $10{,}000$ & $\mathbf{79.5\%}$ & $50.0\%$ & $55.3\%$ \\
\bottomrule
\end{tabular}
\label{tab:b-triplet}
\end{table}

\begin{table}[!b]
\centering
\caption{R@1 accuracy under random vs.\ hard-negative distractor sampling. R@1 varies substantially along both axes: across distractor difficulty (right column, Random$-$Hard) and across set size (bottom row, range over $n$). \textbf{EmCom-Diffusion depends on neither choice and is unaffected by this variation.}}
\label{tab:r1_distractor}
\small
\begin{tabular}{lccc}
\toprule
 & \multicolumn{2}{c}{Distractor sampling} & \\
\cmidrule(lr){2-3}
$n$-way & Random & Hard & Gap \\
\midrule
$32$       & $98.8\%$ & $71.4\%$ & {\boldmath$+27.4$} \\
$100$      & $96.6\%$ & $63.8\%$ & {\boldmath$+32.8$} \\
$1{,}000$  & $79.8\%$ & $55.2\%$ & {\boldmath$+24.6$} \\
$5{,}000$  & $49.8\%$ & $50.8\%$ & {\boldmath$-1.0$}  \\
\midrule
Range      & {\boldmath$49.0$} & {\boldmath$20.6$} & \\
\bottomrule
\end{tabular}
\end{table}

\paragraph{(b) EmCom-Diffusion captures visual distinctions in regimes
where TopSim and CBM cannot.}

TopSim depends on rank correspondence between message distance and image distance, and CBM is bounded by a predefined concept inventory (§\ref{sec:comparison}); both fail when distinct images map to similar token sequences. Small message distance likewise does not imply visual similarity: $81.5\%$ of pairs at message edit distance $1$ are visually dissimilar, $84.8\%$ at $2$, and $96.3\%$ at $3$.
For each anchor $A$ we sample candidates with $d_{\mathrm{msg}}(m_A,m_B) = d_{\mathrm{msg}}(m_A, m_C)$, with $B$ visually similar and $C$ dissimilar. TopSim is operationalized per-pair as $-d_{\mathrm{msg}}(m_A, m_X)$; CBM uses Hungarian bipartite matching between tokens and the MS-COCO concept inventory, scoring pairs by cosine similarity of induced concept multisets.

Table~\ref{tab:b-triplet} reports the results. TopSim is exactly at chance ($50.0\%$) by construction. CBM exceeds chance at small edit distances ($68.8\%$ at $1$) but degrades sharply ($55.3\%$ at $3$) as concept multisets converge with token overlap. EmCom-Diffusion's accuracy is flat across edit distances ($77.7\%$--$79.5\%$).

\paragraph{(c) R@1 depends on distractor choice (§\ref{sec:comparison});
EmCom-Diffusion is invariant.}

We evaluate the same listener and messages under uniform random vs.\ hard-negative distractor sampling, where hard distractors are the $n{-}1$ images with highest CLIP ViT-L/14 cosine similarity to the target. Table~\ref{tab:r1_distractor} reports the results. R@1 under random distractors falls from $98.8\%$ at 32-way to $49.8\%$ at 5{,}000-way (full corpus), while under hard distractors it varies more narrowly. The gap peaks at moderate set sizes (e.g., +32.8 pp at 100-way) and vanishes at 5{,}000-way, where random sampling already draws the hardest negatives and the two schemes coincide. Since the messages and listener are identical across conditions, the gap reflects R@1's dependence on the distractor distribution rather than the visual content of the message. \textbf{EmCom-Diffusion does not depend on distractor selection.}

\section{Conclusion}
\label{sec:conclusion}

We proposed EmCom-Diffusion, which fine-tunes a pretrained text-to-image diffusion model on emergent-language tokens and quantifies visual reflection as the perceptual similarity between the generated and original image, using the input image itself as the standard of comparison and operating generatively.

On MS-COCO, EmCom-Diffusion distinguishes emergent languages from random and fixed baselines across three visual encoders. Compared against four existing metrics, EmCom-Diffusion captures visual content each of them either misses or spuriously credits: supervised translation is bounded by the reference-caption 
distribution and its signal collapses when caption similarity is held constant; TopSim and CBM rely on token-level structure and provide no discriminative signal when visually distinct images receive similar token sequences; Referential Game accuracy R@1 depends on the distractor distribution and degrades substantially under hard distractors despite identical messages and listener. By directly comparing emergent messages with the input images 
they describe, EmCom-Diffusion measures visual reflection in a way unaffected by these limitations.

\paragraph{Limitations.}
Our framework depends on the choice of pretrained image generator: we use Stable Diffusion v1.5, and the inductive biases of the underlying model shape the kinds of visual content it can recover. In practice, even random-token inputs rarely produced completely degenerate images, suggesting that the generator's strong prior contributes a baseline of image plausibility largely independent of the input. Because the generator is fine-tuned on the messages, the score reflects a property of each message together with this fine-tuned model rather than of the message in isolation; the shared random- and fixed-token baselines control for the model's input-independent contribution (§\ref{subsec:validity}). Our analyses are also conducted within a single communication paradigm (the Referential Game) and a single image domain (MS-COCO 2017); whether the framework yields similarly coherent characterizations for emergent languages produced under other game settings, such as description or negotiation, or on datasets with different visual statistics, remains to be tested. Finally, while EmCom-Diffusion distinguishes emergent languages from baselines in aggregate, it does not yet pin down which specific kinds of visual content the tokens preserve and which they discard (e.g., object identity vs. spatial layout vs. texture); resolving this is a natural next step, for example through token-level attribution and finer-grained generation conditioning. 

\paragraph{Future work.}
More broadly, the framework offers a constructive route to the question that motivated this work: how the structure of the physical world comes to be reflected in language. Because emergent communication can be trained from scratch, applying EmCom-Diffusion across training would provide an empirical handle on how such world-reflecting structure forms, complementary to analyses of pretrained models, which observe it only in its mature form.

\section*{AI Usage Declaration}
We used generative AI tools to assist in the implementation of the proposed method and in the preparation of this manuscript. All generated outputs, including source code and text, were reviewed, verified, and approved by the authors, who take full responsibility for the content of this work.

\section*{Appendix}
\setcounter{section}{0}
\renewcommand{\thesection}{\Alph{section}}

\section{Referential Game Setup}
\label{app:speaker}

The speaker is a frozen DINOv2 ViT-B/14~\cite{oquab2024dinov2} encoder followed by a cross-attention module with $K$ learnable queries attending to the patch tokens; the outputs are projected to vocabulary logits to sample $K$ tokens simultaneously via straight-through Gumbel-softmax~\cite{ST,gumbel-softmax}, with temperature annealed from $\tau=1.0$ to $0.5$ over the first $30{,}000$ steps. The listener encodes the message via a $6$-layer Transformer ($6$ heads, hidden size $768$) and projects the pooled output to a $256$-d vector with $L_2$ normalization; each candidate is encoded by the same frozen DINOv2 with a separate projection head, and the highest-cosine-similarity candidate is selected. The two networks are trained jointly on the MS-COCO 2017 training set ($118{,}287$ images) via a $128$-way game with $K=8$, $V=256$, using AdamW~\cite{adamw} (lr $3\times10^{-4}$, weight decay $0.05$, batch size $128$, $1{,}000$-step warmup) for $30$ epochs on $4$ NVIDIA RTX 6000 Ada GPUs (48\,GB each). After training, only the speaker is used to emit one message per image.

\section{Image Generation and Evaluation}
\label{app:training}

For each condition we fine-tune the generator for $10{,}000$ steps (effective batch size $64$; fp16) on the same $4$ GPUs, with AdamW (weight decay $10^{-2}$, gradient clipping $1.0$) and lr $1\times10^{-4}$ for both the emergent token embedding layer (EL) and the U-Net LoRA weights (SD-NL), under a $500$-step warmup then constant schedule. Fine-tuning uses only (image, message) pairs from the MS-COCO 2017 training split; Random and Fixed share EL's setup. We evaluate on a fixed held-out subset of $5{,}000$ validation images (seed $42$), disjoint from training, generating one image per input via DDIM~\cite{ddim} ($50$ steps, guidance scale $7.5$). Per-image similarity uses pretrained CLIP ViT-B/32, DINOv2 ViT-B/14, and SigLIP ViT-B/16.

\section{Comparison Methods: Implementation Details}
\label{app:comparison}

\textbf{CBM}~\cite{carmeli2024concept}. Inventory $\mathcal{C}$: the $80$ MS-COCO categories; per-image dominant category by highest instance count. $q_{vc}$: count of images where token $v$ appears and $c$ is dominant. Matching: Hungarian (\texttt{scipy.optimize.linear\_sum\_assignment}). Per-pair (§\ref{subsec:comparison}~(b)): cosine similarity of the induced length-$80$ category-frequency vectors over the $K{=}8$ tokens.

\textbf{Supervised translation}~\cite{yao2022linking}. $f_\theta$: Transformer encoder-decoder ($3$ encoder, $6$ decoder layers, $d_{\text{model}}=256$) trained on $(m_i, c_i)$ pairs from the MS-COCO 2017 training set (one of five captions per epoch); Adam (lr $3\times10^{-4}$, batch size $256$, $10$ epochs). Corpus-level: ROUGE-L~\cite{lin-och-2004-automaticROUGE-L}. Per-pair (§\ref{subsec:comparison}~(a)): CLIP ViT-L/14 cross-modal cosine between the translation of $m_A$ and the candidate's average GT-caption text embedding.

\textbf{TopSim}~\cite{kirby2006topsim}. $\phi$: DINOv2 ViT-B/14 (CLS output through the listener's \allowbreak \texttt{vision\_proj} head; Appendix~\ref{app:speaker}); message distance: Levenshtein. Corpus-level: Spearman correlation over all $\binom{5{,}000}{2}$ pairs. Per-pair (§\ref{subsec:comparison}~(b)): $-d_{\mathrm{msg}}(m_A, m_X)$.

\textbf{R@1}~\cite{lazaridou2017referential}. Corpus-level: $128$-way matching training (Appendix~\ref{app:speaker}). Distractor analysis (§\ref{subsec:comparison}~(c)): $n \in \{2,4,8,16,32,100,500,1000,5000\}$; hard distractors = the $n{-}1$ images with highest CLIP ViT-L/14 cosine similarity to the target.

%
%
%
\bibliographystyle{splncs04}
\bibliography{bibliography}

\end{document}